\documentclass[10pt]{llncs}
\usepackage[T1]{fontenc}
\usepackage{amsmath,amssymb,amsfonts}
\usepackage{graphicx}
\usepackage{textcomp}
\usepackage{xcolor}
\def\BibTeX{{\rm B\kern-.05em{\sc i\kern-.025em b}\kern-.08em
    T\kern-.1667em\lower.7ex\hbox{E}\kern-.125emX}}
\usepackage[T1]{fontenc}
\usepackage{hyperref}
\usepackage{url}
\usepackage{times}
\usepackage{booktabs}

\usepackage{pbox}
\usepackage{bbm}
\usepackage{subcaption}
\usepackage[rightcaption]{sidecap}
\usepackage{float}
\usepackage{longtable}
\usepackage{multirow}
\usepackage{xcolor}
\usepackage{svg}
\definecolor{commentcolor}{RGB}{110,154,155}

\usepackage{color}

\usepackage{algorithm}
\usepackage{algpseudocode}
\usepackage{array}
\begin{document}
%

\title{Evaluating the Efficacy of Instance Incremental vs. Batch Learning in Delayed Label Environments: An Empirical Study on Tabular Data Streaming for Fraud Detection}
\author{ Kodjo Mawuena Amekoe$^{1,3}$, Mustapha Lebbah$^{1,2}$,Gregoire Jaffre$^{3}$,\\  Hanene Azzag$^{1}$, and Zaineb Chelly Dagdia$^{2}$   
%
%
%
\vspace{.3cm}\\
%
1- Sorbonne Paris Nord University - LIPN, UMR CNRS 7030, Villetaneuse, France
%
\vspace{.1cm}\\
2- Paris-Saclay University - DAVID Lab, UVSQ, Versailles, France 
\vspace{.1cm}\\
3- Groupe BPCE, 7 promenade Germaine Sablon 75013 Paris, France \\
}
\institute{}
\maketitle 
\begin{abstract}
Real-world tabular learning production scenarios typically involve evolving data streams, where data arrives continuously and its distribution may change over time. 
In such a setting, most studies in the literature regarding supervised learning favor the use of instance incremental algorithms due to their ability to adapt to changes in the data distribution. Another significant reason for choosing these algorithms is \textit{avoid storing observations in memory} as commonly done in batch incremental settings. However, the design of instance incremental algorithms often assumes immediate availability of labels, which is an optimistic assumption. In many real-world scenarios, such as fraud detection or credit scoring, labels may be delayed. Consequently, batch incremental algorithms are widely used in many real-world tasks. This raises an important question: "In delayed settings, is instance incremental learning the best option regarding predictive performance and computational efficiency?" Unfortunately, this question has not been studied in depth, probably due to the scarcity of real datasets containing delayed information. 
In this study, we conduct a comprehensive empirical evaluation and analysis of this question using a real-world fraud detection problem and commonly used generated datasets.
Our findings indicate that instance incremental learning is not the superior option, considering on one side state-of-the-art models such as Adaptive Random Forest (ARF) and other side batch learning models such as XGBoost. Additionally, when considering the interpretability of the learning systems, batch incremental solutions tend to be favored.  Code: \url{https://github.com/anselmeamekoe/DelayedLabelStream} 

\end{abstract}
\vspace{-0.5cm}
\section{Introduction}
Currently, monitoring and adapting/updating machine learning models is a clear requirement in many real-world problems. This is mainly due to the changes that may occur in the data distribution over time, violating the identical distribution hypothesis (between the training and production/test dataset) usually made in a classical offline machine learning setting. 
 Additionally, it is impractical to store all data in memory indefinitely in tasks where data is generated at high speed.
Over the last two decades, many efforts have been made to develop learning systems that respect these requirements and are able to provide predictions in real-time. Within these solutions, we can differentiate instance incremental models, which have the capability to update their weights or architecture using just one instance. Most state-of-the-art models in this category are based on the Hoeffding Tree \cite{domingos2000mining} and are often equipped with drift or change detection mechanisms such as ADaptive WINdowing (ADWIN) \cite{bifet2007learning}. We also have, batch incremental learning systems in which the stream datasets are stored in chunks or batches that are further used to learn a new static model or update the old one.
An important consideration in the evolving data stream is label delay \cite{plasse2016handling,grzenda2020delayed,gomes2022survey,haug2022standardized}, where labels can be received with important delays depending on the task. For example, in card fraud detection, the true nature of a card transaction (fraudulent or genuine) is only known after some days or weeks, unless the card owner notices an anomaly on its card and reports it immediately to the bank. In such a situation, it is unavoidable in a supervised paradigm to store observations waiting for their labels; otherwise, semi-supervised or unsupervised systems should be considered \cite{gomes2022survey}. Besides labeling delay and for most real-world problems, it is often a requirement to store observations for some fixed periods. For instance, banks are legally obliged to store transaction information for several months and should make it available upon customer request. 
In this study, we investigate whether instance incremental solutions remain the best option in such delayed situations by considering their predictive performance and computational efficiency.\\
\indent Moreover, the interpretability of automatic decision-making systems is required in many real work cases (e.g., GDPR, Article 22, IA Act in Europe).  Regarding learning with evolving data streams, interpretability involves understanding the decisions made by the model/system, as well as changes within the data distribution and the model itself~\cite{haug2022dynamic}. 
We show that state-of-the-art batch incremental interpretable models outperform instance incremental models in terms of accuracy and explain why they might be preferable for inherent interpretability for evolving data streams.
Overall, we summarize our contributions as follows:
\begin{itemize}
    \item We provide background on delay and interpretability in learning with evolving data streams (Section \ref{sec:relatedwork}).
    \item We designed a realistic supervised evaluation framework based on interleaved chunks, combined with label delays (Section \ref{secEvalproposition}).
    \item We empirically compare instance incremental, and batch incremental algorithms on the designed evaluation framework and revealed the superior performance of the latter in delayed settings using benchmark tabular stream datasets and real-world fraud modeling problem (Section \ref{secExp}).
    \item We analyze the performance of batch incremental models and demonstrate the importance of storing past observations whenever possible, especially for tasks where the target event is rare, such as fraud detection (Section \ref{sec_fraud_result}). 
\end{itemize}
\section{Related work}
\label{sec:relatedwork}
\subsection{Performance evaluation strategy.}
The most common strategy used to assess performance in data streams is the \textbf{prequential} evaluation, also known as the test-then-train strategy. In this approach, each observation is first used for testing (updating the evaluation metric) and then for updating the learning model. Typically, prequential evaluation is combined with a forgetting mechanism~\cite{gama2013evaluating,gama2009issues} (such as a sliding window or fading factor) to place more weight on recent performance. Prequential evaluation is usually applied in settings where labels are available immediately after the observation or after a fixed number of instances. 
Observations may be stored in a buffer or chunk for supervised problems with varying and significant label delays. Once the labels become available, they can be used to test and/or update/retrain the learning model. This strategy is sometimes called \textbf{interleaved chunks}. The interleaved chunks approach allows the use of common batch evaluation metrics, such as AUCROC~\cite{aucroc} and AUCPR~\cite{aucpr}.
\subsection{Batch incremental versus Instance incremental learning.}
The utility of rigorously comparing batch incremental and instance incremental approaches is underestimated in the learning literature on evolving data streams. This is arguably due to the preference for automatically updating the model over time and/or avoiding the storage of observations in memory whenever possible, which favors instance incremental solutions. Consequently, attempts to use batch incremental solutions are often biased towards using \textit{small} predefined chunks/batches/windows of instances to retrain or update an old model automatically.
%
Among these approaches, we cite Accuracy Weighted Ensemble (AWE) \cite{wang2003mining} in which base classifiers (typically C4.5, SVM, Linear models) are weighted based on their expected test accuracy over time. Another example is the Adaptive XGBoost (AXGB) solution \cite{montiel2020adaptive}, where new decision tree models are created and appended to the boosted ensemble.  More precisely, once the ensemble is full, the oldest member is removed before appending a new one (the push strategy), or older members are directly replaced with newer ones (the replacement strategy).\\
\indent Dynamic or continuous offline optimization of batch learners (using data collected over time, e.g., days, months) alternate with online inference (real-time prediction) is often neglected in the literature when comparing batch and instance incremental learning; however, this strategy remains a widely used approach in real-world production/deployment or \textit{Human in Loop} Machine Learning Operations (MLOps) because it makes human inspection and validation (bias correction, understanding of changes) easier, thereby increasing trustworthiness.
%
Among the studies that compare batch incremental and instance incremental learning in a supervised setting, we note the work by Read et al.~\cite{read2012batch}, from which we summarize the main disadvantages of these two approaches in Table \ref{disavantages}.
\begin{table}[t]
    \centering
        \caption{Comparative Disadvantages of Batch and Instance Incremental Learning}
    \label{disavantages}
    \begin{tabular}{|c|m{5.5cm}|}
        \hline
        {Learning Approach} & {Main disadvantages} \\
        \hline
       Batch & 
        \begin{itemize}
            \item Require deciding  the batch/windows size for retraining/updating the model;
            \item Cannot learn the most recent examples until a new batch is complete.
        \end{itemize} \\
        \hline
        Instance & 
        \begin{itemize}
            \item Only learns a concept correctly from a large number of examples (e.g., the convergence of HT to a batch-trained Decision Tree is asymptotic and guaranteed only in a stationary environment)~\cite{domingos2000mining,haug2022dynamic};
            \item Has fewer established results than batch learning (e.g., evaluation, bias correction, interpretability).
        \end{itemize} \\
        \hline
    \end{tabular}
\end{table}

Their experimental result reveals the superior predictive performance (but longer running time) of the instance incremental ensemble Leveraging Bagging with Hoeffding Tree as a base learner (LB-HT) \cite{bifet2010leveraging} over AWE using a window of 500 observations and an ensemble of 10 learners maximum. In \cite{montiel2020adaptive}, the window size was increased to 1,000 and the number of learners to 30 for batch incremental models; however, conclusions remain the same, i.e., the state-of-the-art instance incremental solution namely Adaptive Random Forest (ARF) \cite{gomes2017adaptive} showed an overall best predictive accuracy over both the batch solution AXGB proposed by the authors and the AWE strategy.  An important point missed in the comparison was hyperparameter optimization. Only a fixed reference configuration of hyperparameters was used when comparing models for all benchmark datasets. Can hyperparameter optimization change the final conclusion? The authors in \cite{montiel2020adaptive} demonstrated the influence of optimization steps on the performance of the ensemble of XGBoost (called BXGBoost in the paper), which implies an increase of 14\% in the average accuracy becoming the best batch incremental of their benchmarking. \\
\indent 
In this study, we designed evaluation in accordance with real-world production scenarios and showed that batch incremental solutions such as XGBoost can have a superior predictive ability with a thorough optimization step.
\subsection{Label delay in learning with evolving data stream}
When learning from data streams, there are three main possibilities regarding label availability  \cite{gomes2017adaptive}: (i) immediate, (ii) delayed with a finite time interval, and (iii) delayed indefinitely. There are also settings where all these possibilities can occur simultaneously \cite{gomes2022survey}. 
Most studies in the supervised paradigm assume immediate availability; however, in many real-world tasks, the label may arrive with an important delay. The (finite) delay mechanism can be \textit{deterministic}, meaning fixed for every value of the feature vector, or \textit{(semi) stochastic},  meaning it follows an unknown probability distribution \cite{plasse2016handling}.
An example of a semi-stochastic case is fraud detection, in which the labels of fraudulent transactions are quickly revealed, whereas those of genuine transactions are mostly revealed after a fixed prescription period. The authors of \cite{plasse2016handling} proposed an adaptive Linear Discriminant Analysis classifier combined with a weighing scheme to handle the labeling delay using a real-world credit scoring dataset, but their analysis was limited to only linear models, and no specific evaluation procedure was proposed. In \cite{gomes2017adaptive}, the influence of label delay was investigated using a fixed delay mechanism of 1,000 instances, and the authors concluded that this resulted in an important performance drop for the ARF and Leveraging Bagging models. Unfortunately, the comparison did not include batch incremental algorithms, and while stochastic delay appears to be the most common delay mechanism in real-world problems, it was not investigated.
In \cite{grzenda2020delayed}, the authors designed a delayed evaluation framework, called continuous re-evaluation, where the goal is to assess the system's or algorithm's capability to refine its predictions over time before the delayed label arrives. Although this framework is quite interesting for flight data, it is infeasible for many problems (e.g., fraud detection and online credit scoring) to request the model to make predictions several times for the same instance. Therefore, our evaluation in this study primarily concentrates on the initial (first) prediction request. 
Additionally, the evaluation includes the comparison of instance and batch incremental learning, which, to the best of our knowledge, is still missing in the literature.
\subsection{Interpretability in learning with evolving data stream}
Much effort has been made in model interpretability in static settings, where the typical objective is to understand the decisions made by the underlying model.  For this purpose, two approaches exist: using post hoc tools such as SHAP \cite{SHAP} to explain the decision of black box models or using a
directly inherently interpretable model such as an Explainable Boosting Machine \cite{nori2019interpretml} and TabSRA \cite{TabSRA_ESANN}. Although interpretability in an evolving data stream is still in its early stages, the two approaches used in static settings can also be adapted to stream or online settings. Regarding post hoc solutions, we cite iSAGE \cite{muschalik2023isage} which offers an incremental version to the Shapley Additive Global Importance (SAGE) \cite{SAGE} tool.
Unlike static settings, interpretability also requires understanding changes that occur in the model behaviors or, at least, in its post hoc explanations caused by incremental model updates \cite{haug2022standardized}. Unfortunately, these approaches have to contend with the reliability issue that post hoc methods suffer from in the static settings
\cite{ProblemSHAP}. Therefore, authors such as \cite{haug2022standardized} favor the use of inherently interpretable models for evolving data stream problems, where the more interpretable to humans is likely to change less over time \cite{haug2022dynamic}. Based on this consideration, our experimental results reveal that batch incremental inherently interpretable models, besides being more interpretable (as their architecture or weights change less frequently than their instance incremental counterparts, and therefore easier for human tracking and understanding), demonstrate superior predictive performance compared to their instance incremental counterparts.
\section{Problem formalization}
\label{secEvalproposition}
\subsection{Label delay}
We consider a supervised learning setting for a continuous data stream $S_i  = \{(\mathbf x_i, y_i)\}$ with $i =1,..., T$ where $T \longrightarrow \infty$, $i$ is the unique identifier of each instance. For binary classification $y_i \in \{0,1\}$ or $y_i \in \mathbbm{R}$ for regression tasks. The input feature vector is $\mathbf x = (x_1,...,x_p) \in \mathbbm{R}^p$ (we do not consider settings where the input dimension changes over time).\\ 
\indent In a delayed setting, labels are available with some delay. We denote by $dt(\mathbf x_i)$ the timestamp when observation $\mathbf x_i$ becomes available, $dt(y_i)$ is the timestamp of the corresponding label's arrival, and  $\Delta t(\mathbf x_i) = dt(y_i)-dt(\mathbf x_i)$ is the label delay.
Obviously, $\delta t(y_i)\ge dt(\mathbf x_i)$ i.e., $\Delta t(\mathbf x_i)\ge 0$ and $\Delta t(\mathbf x_i)= 0$ corresponds to the particular case where $y_i$  becomes immediately available (before the observation following $\mathbf x_i$). In general, the units of delay are seconds, minutes, days, and months, but there are also simplified settings where the delay is evaluated in terms of instances \cite{gomes2017adaptive,grzenda2020delayed}.
In such delayed settings, it is unavoidable to store observations until their labels become available.  More precisely, once available, label $y_i$ can be used to update the metric and couple $\{(\mathbf x_i, y_i)\}$ to update the learning model. Following \cite{grzenda2020delayed}, we denote by $S_i = \{(\mathbf x_i, ?)\} $ at the timestamp $dt(\mathbf x_i)$, i.e., when the label is not yet available and $S_i = \{(., y_i)\} $ at the timestamp $dt(y_i)$, i.e., when the label becomes available.
\subsection{Proposed predictive performance evaluation methodology}
Algorithm \ref{alg:batch} summarizes the learning process for batch incremental models.
The inference is made in real-time, i.e., incrementally per instance, as in common stream learning, but the evaluation metric is applied to a chunk or batch of predictions (with potentially varying sizes). The batch is chosen such that there are sufficient labeled observations for consistent performance evaluation (we refer to this $\mathrm{B\_label}[id_{B}].isFull()$ line 11 Algorithm \ref{alg:batch}). For real-world problems, this batch is usually defined in terms of time, for example, one day, one month of predictions, or one year, depending on the speed of the stream and the label delay. For the instance incremental and batch incremental methods, the observations are stored in a buffer B\_X, waiting for their labels, which may be available after some delay. The main difference in data storage between the two learning strategies lies in the updating requirements. For instance incremental learning, every observation $\mathbf x_i$ is stored in the buffer, and once its label $y_i$ becomes available, the pair $(\mathbf x_i,y_i)$ is used immediately to update the model.
For batch incremental learning, observations are stored until there are enough labels for retraining or updating the batch model (line 14 Algorithm \ref{alg:batch}). Therefore, batch incremental systems may require more storage resources than instance incremental ones. However, in situations where delay is significant and variable, the storage requirements can be quite similar, as highlighted in the \textit{fraud dataset} (Section \ref{datasets}).
Without loss of generality, we assume in this study that the number of labeled instances required for retraining or updating the supervised batch models is equal to or greater than the number required for periodic evaluation.\\
\indent Finally, our evaluation framework includes the option of using a pre-trained model at the beginning of the stream for both instance incremental and batch incremental learning.
For many real-world problems, an initial model is typically optimized offline using data collected over time before deploying a learning system. 
Therefore, the stream data represents the starting point of production or deployment. 
This practice is becoming common in the incremental learning literature, where a fraction of data is used either to train an initial model before the stream begins \cite{haug2022dynamic} or for hyperparameter optimization \cite{montiel2020adaptive}.
\begin{algorithm}
{\small
\caption{Evaluation for batch incremental models}\label{alg:batch}
\begin{algorithmic}[1]
\Require $S_1, S_2, S_3$,... - data stream, $f$-initial pretrained model, $eval\_metric$ - evaluation metric e.g., Accuracy, AUCROC, AUCPR.
\Ensure B\_X, B\_label, B\_pred, - buffer for storing observations, labels, predictions respectively
\For{$S_i,i=1,...$}
\State $i \gets get\_index(S_i)$ 
\State $id_{{B}} \gets get\_batch(S_i)$ 
\If{$S_i = \{(\mathbf x_i, ?)\}$} 
\State $y_{pred} \gets f(\mathbf x_i)$ \Comment{Get the real-time prediction}
\State $\mathrm{B\_X}[id_{{B}}][i].add(\mathbf x_i)$ \Comment{Add the observation to the buffer}
\State $\mathrm{B\_pred}[id_{{B}}][i].add(y_{pred})$ 
\ElsIf{$S_i = \{(., y_i)\}$} 
\State $\mathrm{B\_label}[id_{{B}}][i].add(y_i)$ 
\EndIf 
\If{$\mathrm{B\_label}[id_{{B}}].isFull()$}
\State $ \mathrm{result} \gets eval\_metric( \mathrm{B\_label}[id_{{B}}], \mathrm{B\_pred}[id_{{B}}])$
\State $display(id_{{B}}, \mathrm{result})$
\If{$\mathrm{B\_label}.isFullForTrain()$} 
\State $i_{train},labels\gets \mathrm{Y\_label}.get\_train\_batch() $  
\State $ X \gets \mathrm{B\_X}.get\_batch\_data(i_{train})$
\State $train(f, \{X, labels\})$ \Comment{Update the model}
\State $update(\mathrm{B\_X})$ \Comment{update the buffer by deleting unnecessary observations}
\State $update(\mathrm{B\_label})$ 
\State $update(\mathrm{B\_pred})$ 
\EndIf
\EndIf 
\EndFor
\end{algorithmic}
}
\end{algorithm}

\section{Experiment analysis}
\label{secExp}
\subsection{Experiment setup}
\subsubsection{Datasets}
\label{datasets}
We consider two types of datasets: (1) benchmark datasets for generated binary classification, well-known in the literature on evolving data streams \cite{gomes2017adaptive,montiel2020adaptive,gunasekara2024gradient,haug2022dynamic}, denoted as {{generated benchmark}}, and (2) a real-world bank transfer {{Fraud}} detection dataset.

\textbf{\textit{Generated benchmark}}.
The datasets in this benchmark are publicly available and are summarized in Table \ref{StatDatasets}: 
\begin{itemize}
    \item \textbf{AGR$_a$} \cite{gomes2017adaptive,montiel2020adaptive,gunasekara2024gradient}: This dataset includes six discrete features and three continuous features. Instances are categorized into two classes using various functions, some of which adhere to decision rules, making it conducive to decision tree analysis. An abrupt drift occurs after every 250,000 instances (three in total).
    \item \textbf{AGR$_g$} \cite{gomes2017adaptive,montiel2020adaptive,gunasekara2024gradient}: Similar to  AGR$_a$. Here, a gradual drift is used.
    \item \textbf{HYPER$_f$} \cite{gomes2017adaptive,montiel2020adaptive,haug2022dynamic}: Simulates an incremental (fast) drift by changing the equation of hyperplane separating the two classes over time. This dataset is, therefore, linear model friendly. The number of features is set to 10.
    \item \textbf{SEA$_a$}\cite{gomes2017adaptive,montiel2020adaptive,haug2022dynamic}:
    It comprises three continuous features $(x_1,x_2,x_3)$, with only the first two being pertinent to the target class. The first two dimensions are divided into four blocks. Within each block, an instance is assigned to class 1 if $x_1 + x_2 \leq \theta$, and to class 0 otherwise, with $\theta$ taking values from the set ${8, 9, 7, 9.5}$. Additionally, an abrupt drift occurs after every 250,000 instances (three in total).    
    \item \textbf{SEA$_g$} \cite{gomes2017adaptive,montiel2020adaptive}: Similar to  SEA$_a$. Here, a gradual drift is employed rather than an abrupt one.

\end{itemize}
\begin{table}[t]
\begin{center}
\caption{Datasets. \textbf{$N$}: number of instances, \textbf{$p$}: number of features, Type: \textbf{S}ynthetic and \textbf{R}eal. Drift \textbf{A}: abrupt, \textbf{G}: gradual, \textbf{I$_f$}: incremental fast, and ?: unknown, MC: Minority Class} 
\label{StatDatasets}
    \begin{tabular}{c c c c c c c}
    \hline
     Datasets & $N$ & $p$ & $\#$  classes & Type & Drift & MC (\%) \\ 
     \hline
     AGR$_a$  & 1M & 9 & 2 & S & A & 47.17\\
      \hline 
      AGR$_g$ & 1M & 9 & 2 & S & G & 47.17\\
     \hline 
      HYPER$_f$ & 1M & 10 & 2 & S &   I$_f$ & 50. 00\\
      \hline 
      SEA$_a$ & 1M & 3 & 2 & S & A & 40.09\\
      \hline
      SEA$_g$ & 1M & 3 & 2 & S & G & 40.09\\ 
      \hline
     Fraud & $\sim$ 6.5 M & 18 & 2 & R & ? & <0.10 \\ 
      \hline
    \end{tabular}
\end{center}
\end{table}

\textbf{Induce label delay in the generated benchmark.}
First, 10\% of each dataset (100,000 instances) is reserved as offline collected data, used for hyperparameter tuning \cite{montiel2020adaptive} and pretraining an initial model \cite{haug2022dynamic} in both instance incremental and batch incremental learning.
Therefore, 90\% of each dataset is used for stream (online) evaluation. Each evaluation batch (Algorithm \ref{alg:batch}) contains approximately 1\% of the dataset. Specifically, the evaluation batch size is generated following a Poisson distribution with a mean of 10,000.
This varying evaluation batch size effectively reflects real-world settings, where monitoring logs of deployed learning systems must be displayed periodically (every day, week, or year, depending on the stream's speed). However, the number of instances may vary from one period to another.\\
\indent In this study, we generated the label delay by assuming it is stochastic and follows a Poisson distribution.
More precisely, we assume that the delay $\Delta t(\mathbf{x_i})$ for each instance observation $\mathbf{x_i}$ follows a Poisson distribution with a mean of $\alpha \times 10,000$ instances, where $\alpha \in \{0, 0.1, 1, 2, 3, 4, 5, 6, 7\}$.\\
\indent \textbf{Fraud dataset}.
The task consists in using known information about fraudulent bank transfer scenarios (via supervised machine learning) to assign a fraud score to each newly added IBAN (International Bank Account Number). Specifically, the goal is to predict the probability that a new IBAN will be used for fraudulent transfers within the next 30 days.
The motivation for investigating instance incremental learning is the variation in label delay based on class labels. Most fraudulent IBANs are identified within a few days, while the remaining IBANs are automatically labeled as genuine after 30 days. 
Hence, one might question whether quickly adapting the learning model with these fraudulent instances could yield better performance compared to a batch learning strategy, where the model is updated only after collecting labels for at least a month. For example, in batch learning, labels from September are collected by the end of October, and the model is updated in early November, whereas instance incremental learning updates the model with September's fraudulent data as soon as it becomes available.
A key challenge is the class imbalance, where only about 0.10\% of instances are fraudulent on average. As a result, it is necessary to store at least 99.90\% of instances for 30 days due to label delay. Additionally, unlike datasets in the \textit{Generated benchmark}, this dataset may experience various types of drift, such as gradual drift combining abrupt changes, and some concepts may reoccur over time.
The sample used in this study comprises approximately 6.5 million observations collected from September 2021 to August 2022.
The initial three months (September, October, and November) are reserved as offline collection for hyperparameter optimization and initial model training. The stream evaluation spans eight months (from January 2022 to August 2022), with each month comprising approximately 540,000 instances.
\subsubsection{Models benchmarked}
\label{models}
For instance incremental learning, we consider:
\begin{itemize}
    \item \textbf{Adaptive Random Forest (ARF)} \cite{gomes2017adaptive}: This model is one of the best performing among instance incremental algorithms \cite{gomes2017adaptive,montiel2020adaptive,gunasekara2024gradient}. We use ARF with the ADWIN detector \cite{bifet2007learning}.
    \item \textbf{Leveraging Bagging with Hoeffding Tree as base learner (LB\_HT)}  \cite{bifet2010leveraging}: We include this model in our comparison as it is the best performing model in \cite{read2012batch}.
    \item \textbf{Leveraging Bagging with Logistic Regression as base learner (LB\_LR)}  \cite{bifet2010leveraging}: We explore in this study a non tree base learner for Leveraging Bagging. More specifically we use the Logistic Regression (LR) with a Stochastic Gradient Optimizer (SGD).
\end{itemize}
Alongside state-of-the-art instance incremental models, we include the following glass box algorithms for interpretability, as discussed in Section \ref{sec:relatedwork}:
\begin{itemize}
    \item \textbf{Logistic Regression (LR)}: This model can be used in both batch or instance incremental models. In this study, we use it as an instance incremental model by combining it with a Stochastic Gradient Optimizer (SGD)
    \item \textbf{Hoeffding Tree (HT)} \cite{domingos2000mining}:
    This model offers interpretable and adaptive decision rules by following paths in the tree, particularly if it's shallow. Thus, we cap the maximum depth at 6. Additionally, we employ Naive Bayes Adaptive for leaf prediction, which enhances the performance of the traditional HT \cite{haug2022dynamic,gama2003accurate}.
    \item \textbf{Hoeffding Adaptive Tree (HAT)}: The same model as the Hoeffding Tree, but with ADWIN \cite{bifet2007learning} for drift handling.
\end{itemize}
For all instance incremental models, we used the Python-based package River \cite{montiel2021river}.
As batch incremental models, we consider:
\begin{itemize}
    \item \textbf{XGBoost} \cite{Xgboost}: 
    It is a leading model in batch learning across various real-world applications and tabular learning competitions. We make XGBoost adaptive by retraining it from scratch after every new batch (Algorithm \ref{alg:batch}, line 17), referred to as \textbf{r\_XGBoost}. Following \cite{montiel2020adaptive,dal2014learned}, we use a stacking of the last $M=3$ retrained XGBoost models, denoted as \textbf{B\_XGBoost}. Although stacking may increase prediction time, it helps preserve some previously learned concepts. We don't include Adaptive XGBoost (AXGB) 
    as it was already outperformed by the optimized version in \textbf{B\_XGBoost}. 
    \item \textbf{Explainable Boosting Machine (EBM)} \cite{nori2019interpretml}: 
    EBM is a Generalized Additive Model with pairwise interaction terms, featuring shape functions that are piece-wise constant and optimized using a boosting mechanism. We include this model in our comparison for its inherent interpretability—its shape function can be monitored to detect changes over time—unlike XGBoost.
    Similarly to XGBoost, we used retraining from scratch (\textbf{r\_EBM}) and stacking (\textbf{B\_EBM}) to make the EBM model adaptive over time. Rather than stacking naively ($M=3$) EBM models, which may compromise interpretability, we merged the shape functions\footnote{\url{https://interpret.ml/docs/python/examples/merge-ebms.html}} of the stacked models into a single final EBM model. 
    \item \textbf{TabSRA} \cite{TabSRA_ESANN}: It is also an inherently interpretable model based on an attention mechanism called self-reinforcement . Monitoring TabSRA's attention weights and inherent feature attributions can help to understand changes over time. We also used the retraining (\textbf{r\_TabSRA}) to make the TabSRA model adaptive. In addition, we investigate the continual fine-tuning of the weights of an initially trained TabSRA model. The learning rate is reduced over time to avoid a catastrophic forgetting of previously learned concepts. This strategy is named \textbf{u\_TabSRA}.
\end{itemize}

\subsubsection{Experiment details on the generated benchmark}
\label{expe_details_generates_benchmark}
\indent \textbf{\\Tuning step.}
The experiments on the \textit{Generated benchmark} are done on 64-Core Processor CPU machine.
For both instance and batch incremental models, hyperparameter optimization is conducted for each dataset using 10\% of the data (considered as offline collected data). Specifically, 30 trials of Bayesian optimization (using Optuna \cite{optuna_2019}) are employed, with a maximum search time of 6 hours for each model and dataset. The best validation model (using prequential evaluation for instance incremental models and a 70/30 train/validation split for batch incremental models) is initialized at the stream's outset and updated over time for instance incremental models. To ensure a fair comparison, hyperparameter optimization is not conducted during the stream for batch incremental models; instead, the best hyperparameter configuration obtained from offline optimization is used for model retraining and updating, as outlined in \cite{montiel2020adaptive}.
Furthermore, {only 70\% of the batch is used for training during the data stream} (with 30\% reserved, as commonly practiced in batch learning, for validation, post hoc analysis, early stopping, bias correction, etc.).
For simplicity, the evaluation batch in the \textit{Generated benchmark} matches the training batch (Algorithm \ref{alg:batch}, lines 11 and 14)\footnote{In this specific scenario, our proposed evaluation methodology aligns with interleaved chunks (Section \ref{sec:relatedwork}) for batch incremental models.}.

For the \textit{Fraud} dataset, we examined a scenario where the evaluation batch spans one month while the training batch covers three months (Section \ref{sec_fraud_result}).
Finally, for all ensemble models (ARF, LB, XGBoost), the maximum number of learners is set to 100, and for tree-based models (ARF, LB\_HT, HAT, HT), the maximum depth is limited to 6 (please check GitHub\footnote{\url{https://github.com/anselmeamekoe/DelayedLabelStream}} for information regarding the hyperparameter search space).\\
\indent \textbf{Evaluation metric and results aggregation.}
For numerous binary classification tasks, such as fraud detection and online credit scoring, practitioners often seek the classifier's confidence output rather than just the binary class obtained by applying a threshold. Therefore, we suggest using AUCROC \cite{aucroc} as the evaluation metric for each stream batch. To summarize the results across all batches (90 in total), we calculate the average and standard deviation of the AUCROC values. Additionally, we include the running time as a metric to assess the computational efficiency of each algorithm.\\
\indent We point out that we do not use cross-validation in our evaluation process, as we cannot alter observations' order without introducing artificial drift \cite{haug2022dynamic}. There are approaches of using different seeds in model training if applicable \cite{gunasekara2024gradient} or using distributed evaluation \cite{bifet2015efficient}. However, they are very computationally intensive, especially when adding hyperparameters optimization steps. 
For more details about the statistical significance in evaluating data streams, we refer the interested readers to \cite{bifet2015efficient,haug2022standardized,haug2022dynamic}.\\
We analyze in the next section the results on the \textit{Generated benchmark}, which contains homogeneous datasets with well-known drift types and difficulties.
\subsection{Results on the generated benchmark}
\begin{table*}[t]
\begin{center}
\caption{Predictive performance on the \textit{generated benchmark} in \textbf{no delay setting}. Mean and standard deviation AUCROC (\%), reported for a data stream split in 90 batches. Italic highlights the best performance when comparing inherently interpretable (II) models, and bold is used for the overall best-performing model. We also report the average (resp. the average normalized by the best AUCROC of the given dataset)  across the 5 stream datasets denoted as Avg (resp. N\_Avg), as well as the average rank (Avg\_Rank). Type: \textbf{II}=Inherently interpretable, \textbf{FC}= Full complexity model}
\label{tab:delay0}
\begin{tabular}{l|r|r|r|r|r|r|rrr}
\hline
Model & Type &AGR$_a$ & AGR$_g$ & HYPER$_f$ & SEA$_a$ & SEA$_g$ & Avg &  N\_Avg& Avg\_Rank \\
\midrule
LB\_HT &FC& 94.54 \tiny{$\pm$4.43}& 90.96 \tiny{$\pm$6.74}& 93.41 \tiny{$\pm$0.39}& 88.90 \tiny{$\pm$0.89}& 88.87 \tiny{$\pm$0.83}& 91.34 & 98.01 & 6.20 \\
LB\_LR &FC& 56.40 \tiny{$\pm$5.94}& 55.92 \tiny{$\pm$5.59}& 94.80 \tiny{$\pm$0.23}& 88.83 \tiny{$\pm$0.88}& 88.78 \tiny{$\pm$0.84}& 76.94 & 83.14 & 9.00 \\
ARF &FC& 94.82 \tiny{$\pm$4.12}& 92.12 \tiny{$\pm$6.12} & 90.36 \tiny{$\pm$1.29}& {88.92} \tiny{$\pm$0.89}& {88.89} \tiny{$\pm$0.84}& 91.02 & 97.67 & 6.00 \\
r\_XGBoost&FC & \textbf{97.36} \tiny{$\pm$8.16}& \textbf{96.25} \tiny{$\pm$4.80}& 92.89 \tiny{$\pm$0.98}& 88.80 \tiny{$\pm$0.86}& 88.75 \tiny{$\pm$0.87}& \textbf{92.81} & \textbf{99.52} & 5.40 \\
B\_XGBoost&FC & 96.59 \tiny{$\pm$9.13}& 95.91 \tiny{$\pm$5.72}& 92.27 \tiny{$\pm$1.40}& 88.89 \tiny{$\pm$0.88}& 88.84 \tiny{$\pm$0.86
}& 92.50 & 99.21 & 5.80 \\
HT &II& 86.50 \tiny{$\pm$12.52}& 79.77 \tiny{$\pm$13.69}& 65.80 \tiny{$\pm$15.17}& 88.28 \tiny{$\pm$0.50}& 88.24 \tiny{$\pm$0.49}& 81.72 & 87.92 & 11.60 \\
HAT  &II& 93.86 \tiny{$\pm$4.98}& 89.72 \tiny{$\pm$7.23}& 89.98 \tiny{$\pm$1.52}& 88.38 \tiny{$\pm$0.90}& 88.21 \tiny{$\pm$0.75}& 90.03 & 96.62 & 10.80 \\
LR  &II& 58.34 \tiny{$\pm$7.49}& 58.12 \tiny{$\pm$6.98}& \textbf{\textit{94.85}} \tiny{$\pm$0.23}& {88.90} \tiny{$\pm$0.88}& 88.86 \tiny{$\pm$0.84}& 77.82 & 84.05 & 7.00 \\
DT  &II& 96.12 \tiny{$\pm$7.70}& 95.09 \tiny{$\pm$4.90}& 86.52 \tiny{$\pm$1.86}& 87.87 \tiny{$\pm$1.29}& 87.74 \tiny{$\pm$0.89}& 90.67 & 97.25 & 9.80 \\
r\_EBM  &II& 95.84 \tiny{$\pm$8.29}& 95.21 \tiny{$\pm$4.86}& 93.43 \tiny{$\pm$0.97}& 88.91 \tiny{$\pm$0.90}& 88.87 \tiny{$\pm$0.86}& 92.45 & 99.16 & 4.40 \\
B\_EBM  &II& 95.49 \tiny{$\pm$9.49}& 94.84 \tiny{$\pm$5.69}& 92.41 \tiny{$\pm$1.90}& \textbf {\textit{88.94}} \tiny{$\pm$0.90}& \textbf{\textit{88.90}} \tiny{$\pm$0.86}& 92.12 & 98.81 & 4.80 \\
r\_TabSRA  &II& 97.10 \tiny{$\pm$8.23}& 95.73 \tiny{$\pm$5.23}& 93.07 \tiny{$\pm$1.20}& 88.89 \tiny{$\pm$0.91}& 88.88 \tiny{$\pm$0.86}& \textit{92.73} & \textit{99.45} & \textbf{\textit{4.20}} \\
u\_TabSRA  &II& \textit{97.16} \tiny{$\pm$7.79}& \textit{95.75} \tiny{$\pm$5.22}& 92.78 \tiny{$\pm$1.20}& 88.74 \tiny{$\pm$1.30}& 88.83 \tiny{$\pm$0.85}& 92.65 & 99.36 & 6.00 \\
\bottomrule
\end{tabular}
\end{center}
\end{table*}
%
\subsubsection{Do instance incremental algorithms really outperform batch incremental ones?}
\begin{figure*}[t]
    \begin{subfigure}{0.45\textwidth}
        \includegraphics[width = 0.99\textwidth,height=4cm]{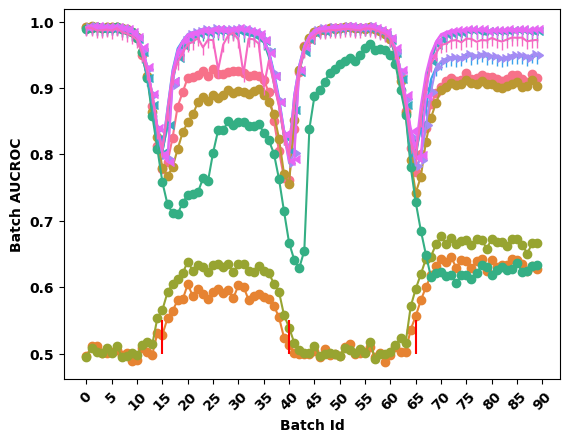}
        \caption{ AGR$_g$: gradual drift (no delay)}
        \label{fig:AGR_g_delay0}
    \end{subfigure}  
    \hspace{0.1cm}
    \begin{subfigure}{0.54\textwidth}
        \includegraphics[width = 0.99\textwidth,height=4cm]{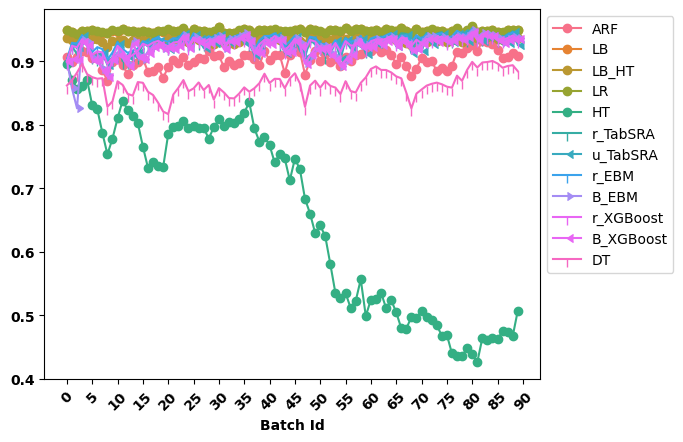}
        \caption{ HYPER$_f$: incremental fast (no delay)}
         \label{fig:HYPER_f_delay0}
    \end{subfigure}
    \begin{subfigure}{0.45\textwidth}
        \includegraphics[width = 0.99\textwidth,height=4cm]{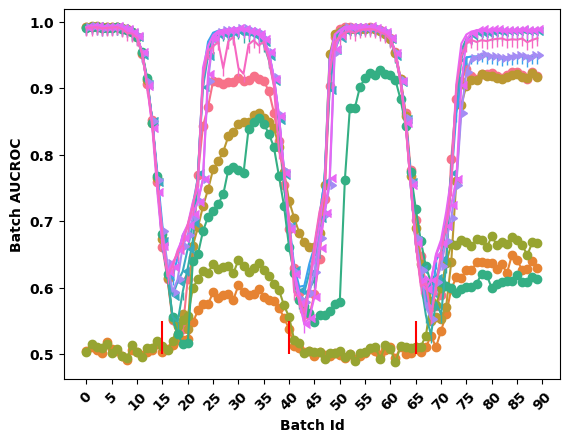}
        \caption{ AGR$_g$: gradual drift (delay factor=7)}
        \label{fig:AGR_g_delay70000}       
    \end{subfigure}
    \hspace{0.1cm}
    \begin{subfigure}{0.54\textwidth}
        \includegraphics[width = 0.99\textwidth,height=4cm]{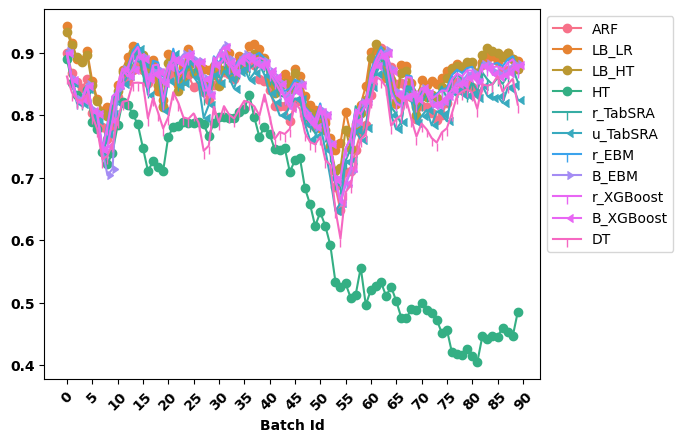}
        \caption{ HYPER$_f$: incremental fast (delay factor=7)}
         \label{fig:HYPER_f_delay70000}
    \end{subfigure}     
    \caption{AUCROC over time. The x-axis is the identifier (ID) of the batch/chunk, which is up to 90. The delay factor 0 corresponds to the no-delay setting, and 7 corresponds to the Poisson stochastic delay of average of 70,000 instances.  The red vertical line indicates the change point for the SEA dataset Fig. \ref{fig:AGR_g_delay0} and \ref{fig:AGR_g_delay70000}.}
    \label{fig:delay0}
\end{figure*}

We analyze the predictive performance (Table \ref{tab:delay0} and Fig. \ref{fig:delay0}) by considering first a no delay setting \cite{read2012batch,gomes2017adaptive,montiel2020adaptive,gunasekara2024gradient}, however using an initial tuning step for both instance and batch incremental as described in Section \ref{expe_details_generates_benchmark} in contrast to previous work.
The key findings are:
\begin{itemize}
    \item With an initial tuning step, the best batch incremental system with an \textit{appropriate} learning batch often achieves a comparable or even superior performance to instance incremental systems for known unique type of drift (including abrupt, gradual, incremental).
    \item When considering interpretable solutions, batch incremental models (such as EBM and TabSRA) offer the advantage of less frequent changes in weights and architecture, making them easier for humans to follow. In addition, these models generally outperform instance incremental models in terms of predictive performance.
\end{itemize}
An important observation regarding the \textit{Generated benchmark} is the impact of the inductive biases of the learning model. Specifically, linear-based models like LR and LB\_LR perform poorly on piece-wise constant-like datasets (AGR$_a$, AGR$_g$). Conversely, tree-based models (DT, HAT, HT) are less effective for modeling continuous (linear) data streams like HYPER$_f$, although ensembling mitigates this issue (for XGBoost, EBM, LB\_HT, ARF).
Hence, we emphasize the importance of alerting readers/practitioners that relying solely on average performance measures from \textit{synthetic benchmark}, which encompasses varying (unequal) types of bias/difficulty, to select stream learning approaches may be misleading for real-world problems.\\
\indent Regarding the adaptation over time, the Hoeffding Tree (HT) appears to be more affected by changes (Fig. \ref{fig:AGR_g_delay0} and \ref{fig:HYPER_f_delay0}) arguably due to its lack of explicit drift handling compared to the HAT (Hoeffding Adaptive Tree). We recall that the maximum depth of the tree is capped to 6 (for DT, HT, HAT) during the hyperparameters optimization; otherwise, it would be challenging to consider them as inherently interpretable. Overall, we observe that batch-based solutions (whether inherently interpretable or black box) can adapt to changes through incremental retraining, even without explicit drift management.\\
\indent An important observation from this experimentation is the hyperparameter sensitivity of tree-based instance incremental algorithms.
As long as the concept\footnote{The function 5 of the Agrawal generator.} on which the parameters were tuned is still valid, instance incremental models with good inductive biases (tree-based ones such as ARF, LB\_HT) manage to match the performance of batch incremental counterparts (highlighted by the Fig. \ref{fig:AGR_g_delay0} for AGR$_g$ dataset with Batch Id $\in [0,15]$). However after the change or drift (Batch Id $\in [15,40]$) these models struggle to converge, unlike batch learners. Yet, once the initial concept reoccurs (Batch Id $\in ]40,65[$) the the models (instance incremental as well) converge to perfect AUCROC again.\\
\indent 
Finally, we observe that for this \textit{Generated benchmark}, employing stacking/updating of models learned over different periods (B\_XGBoost, B\_EBM, u\_TabSRA) does not yield improved predictive performance; instead, it increases computational resources or running time (Table \ref{tab:runtime}). We argue that this is due to the simplicity of the datasets, where typically small learning datasets (roughly 10,000 instances) suffice to learn the underlying concept. However, for more complex tasks, we anticipate that stacking may prove beneficial, as demonstrated in Section \ref{sec_fraud_result} for the \textbf{\textit{Fraud}} dataset.
\subsubsection{How does the label delay impact incremental algorithms?}
After comparing algorithms in no delay settings, it is crucial to consider the delay setting, which involves answering the following questions: (1) How does the label delay impact the performance of the stream learning system? (2) is instance incremental learning the best option regarding the predictive performance in such delayed settings?
\begin{figure*}[t]
    \begin{subfigure}{0.49\textwidth}
        \includegraphics[width = 0.99\textwidth,height=3.3cm]{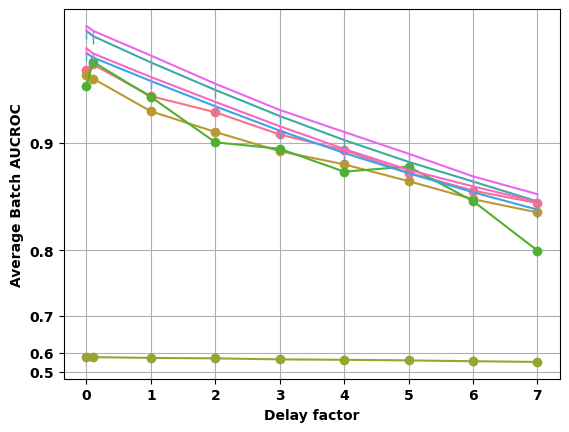}
        \caption{ AGR$_a$: abrupt drift}
                \label{fig:AGR_a_DELAY_Influence}
    \end{subfigure}
    \begin{subfigure}{0.49\textwidth}
        \includegraphics[width = 0.99\textwidth,height=3.3cm]{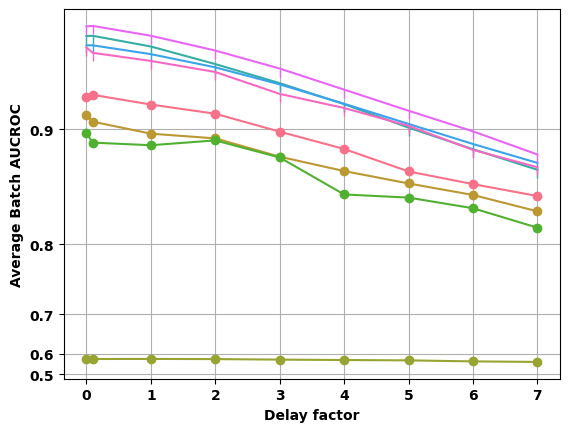}
        \caption{ AGR$_g$: gradual drift}
        \label{fig:AGR_g_DELAY_Influence}
    \end{subfigure}
    \begin{subfigure}{0.49\textwidth}
        \includegraphics[width = 0.99\textwidth,height=3.3cm]{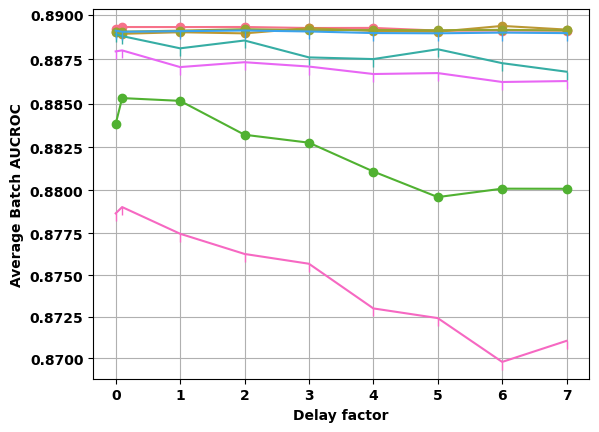}
        \caption{ SEA$_a$: abrupt drift}
        \label{fig:SEA_a_DELAY_Influence}
    \end{subfigure} 
    \begin{subfigure}{0.51\textwidth}
        \includegraphics[width = 0.99\textwidth,height=3.3cm]{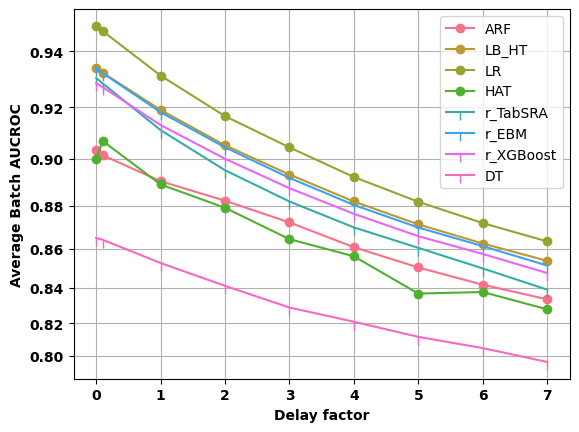}
        \caption{ HYPER$_f$: I$_{f}$}
        \label{fig:HYPER_f_DELAY_Influence}
    \end{subfigure}    
    \caption{Impact of the label delay on the predictive performance. The x-axis indicates the delay factor ranging from 0 (no delay) to 7 (the Poisson stochastic delay of average 70,000 instances). The best model for each category is reported to make the figures more clearer (LB\_LR, HT, u\_TabSRA, B\_EBM, B\_EBM) as discarded. I$_{f}$ = incremental fast drift.}
    \label{fig:DELAY_Influence}
\end{figure*}
From Fig. \ref{fig:delay0}, we can notice that the labeling delay impacts the recovery or adaptation of both instance and batch learners to changes, consequently affecting overall performance (Fig. \ref{fig:DELAY_Influence}), consistent with findings \cite{gomes2017adaptive,gomes2022survey}. However, this impact tends to be limited for low-severity drift/changes. For the SEA$_{a}$ (Fig. \ref{fig:SEA_a_DELAY_Influence}) dataset for instance (where the changes in the simulated decision functions are not significant), the average drop in AUCROC in delayed context seems very low, even negative for some models (for both gradual and abrupt drift) in contrast to others generated datasets (AGR$_g$, AGR$_a$, HYPER$_f$).\\
\indent Besides the impact of the labeling delay on all stream learning models, we can notice (Fig. \ref{fig:DELAY_Influence}) that the best-performing batch learner usually provides similar or superior results compared to instance counterpart regardless of the length of the delay.\\
\subsubsection{Computational efficiency}
In this part, runtimes of the compared models during the stream are analyzed, encompassing retraining and inference, as a measure of computational efficiency. We recognize that relying solely on runtime may not suffice, as it can vary with implementation. For tree-based models, parameter or node count is sometimes considered \cite{montiel2020adaptive}, yet comparing computational efficiency across different model classes (e.g., neural networks versus tree-based models) using this metric is challenging.\\
\indent
As shown in Table \ref{tab:runtime}, state-of-the-art instance incremental learners like ARF are over twice as costly as batch counterparts such as r\_XGBoost. We argue that this is due to the explicit drift monitoring process employed in instance-based learning models. Furthermore, employing batch model stacking (B\_XGBoost, B\_EBM) tends to notably increase inference time and, consequently, the runtime of the stream.\\
\indent
Concerning inherently interpretable solutions, batch incremental options like EBM and TabSRA exhibit longer runtimes compared to instance incremental models (HT, HAT, LR). It is worth noting that the significant runtime difference between Decision Tree (DT) and shallow instance incremental learners (HT, HAT, LR) is arguably attributed to the data filtering process required for selecting new batches for model retraining/updating (Ligne 15 Algorithm \ref{alg:batch}).
   
\begin{table}[t]
\begin{center}
\caption{Evaluation time (s). The average (Avg.) and standard deviation (Std.) time (the smaller, the better) is reported from 9 delayed configurations (factor). }
\label{tab:runtime}
\begin{tabular}{llrr}
\toprule
 Model & Avg. runtime & Std. runtime  \\
\midrule
LB\_HT & 5837.58 & 4428.63 \\
LB\_LR & 7279.24 & 4273.45 \\
ARF & 3702.08 & 2622.41 \\
r\_XGBoost & 1688.97 & 210.53 \\
B\_XGBoost & 5101.62 & 1186.93 \\
\hline
HT & 60.07 & 11.90 \\
HAT & 248.20 & 63.00 \\
LR & 83.93 & 29.07 \\
DT & 976.94 & 47.11 \\
r\_EBM & 1456.71 & 403.83 \\
B\_EBM & 4167.64 & 3244.12 \\
r\_TabSRA & 2117.04 & 298.78 \\
u\_TabSRA & 2685.77 & 166.45 \\
\bottomrule
\end{tabular}
\end{center}
\vspace{-5mm}
\end{table}
\subsection{Results on the Fraud dataset}
\label{sec_fraud_result}
In this section, we compare the models using the \textit{Fraud dataset} (described in Section \ref{datasets}). Unlike the \textit{Generated benchmark}, this dataset displays a significant class imbalance (Table \ref{StatDatasets}), a common situation in real-world applications such as fraud detection. For this dataset, we also consider both instance and batch incremental models presented in Section \ref{models} except LB\_LR which tends to be particularly ineffective at modeling non-linear functions and has high runtime demands (Table \ref{tab:runtime}). Furthermore, we incorporate two learning strategies for batch learners:
\begin{itemize}
    \item \textbf{Static}: All batch learners are optimized using the initial offline collected datasets of three months but are not retrained/updated during the stream. This approach saves from retraining and validation costs, but may lead to significant performance drop after severe changes or drifts. Models trained with the static strategy will be prefixed with \textbf{static\_}.
    \item \textbf{Propagate}: The instance from past fixed number of chunks (months in our case) are propagated and combined with the currently available one for retraining/updating the batch model. This approach may be very beneficial for tasks where certain groups or categories are very underrepresented or target event is very rare \cite{dal2014learned}. However, this approach may add an additional cost for storing past batches and may impair the adaptation to the newest concept, especially when the number of past-considered chunks is important. In this study, we explore the trade-off by investigating the combination of the two previous months (chunks) with the current one. This situation is equivalent to the case where the learning chunk is three times the evaluation batch (Algorithm \ref{alg:batch}). Models trained with this strategy will be prefixed with \textbf{propagate\_}.  
\end{itemize}
Similarly to the \textit{Generated benchmark}, we use an initial optimization step for the \textit{Fraud} dataset, as highlighted in Section \ref{datasets}. We optimize the batch incremental algorithm using 10 different parameter configurations and the instance incremental algorithms using 15 configurations\footnote{The experiment on this dataset is conducted on a private business Google Cloud Platform provided by our data provider, where running very long experiments is infeasible. Therefore, we used a maximum of 15 iterations (instead of 30 for the generated datasets) as a trade-off between tunability and predictive accuracy} including resampling methods \cite{ferreira2019adaptive,aguiar2022survey}. We consider undersampling for ARF and LB\_HT, which already have a long runtime due to online bagging, and investigate undersampling for HT and HAT. For LR, we examine cost sensitivity by modifying the misclassification cost of positive examples in the hyperparameter configuration\footnote{\url{https://riverml.xyz/0.9.0/examples/imbalanced-learning/}}. The performance of instance incremental models was particularly poor without explicitly handling class imbalance.
{
\begin{table*}[t]
\begin{center}
\caption{Predictive performance on the fraud dataset. The evaluation is done in batches of one month, starting from January 2022 to August 2022, while predictions are made in real time (one instance at a time). AUCPR is used as a metric, and we also report the Average (Avg) and Standard deviation (Std) over the 8 months. Italic highlights the best performance when comparing inherently interpretable models (HT, HAT, LR, DT, EBM, TabSRA) models, and bold is used for the overall best-performing model.}
\label{tab:fraudataset}
\begin{tabular}{l|p{.85cm}|p{.85cm}|p{.85cm}|p{.85cm}|p{.85cm}|p{.85cm}|p{.85cm}|p{.85cm}|p{.85cm}|p{.85cm}|}
\toprule
Model &     Jan &     Fev &     Mar &     Apr &     May &     Jun &     Jul &     Aug &     Avg &    Std \\
\midrule
LB\_HT             &  5.26 &  7.32 & 10.89 &  6.76 & 10.18 &  8.27 & 10.65 & 19.27 &  9.83 & 4.03 \\
ARF               &  9.38 & 14.08 & 22.73 & 13.12 & 12.50 & 12.27 &  9.90 & 19.87 & 14.23 & 4.39 \\
static\_XGBoost    & \textbf{15.68} & 23.12 & \textbf{33.76} & 14.69 & 19.48 & 17.31 & 12.75 & 22.55 & 19.92 & 6.25 \\
r\_XGBoost         & \textbf{15.68} & 23.00 & 24.96 & 17.53 & 18.23 & 20.54 & 14.65 & 18.60 & 19.15 & 3.29 \\
B\_XGBoost         & \textbf{15.68} & 23.00 & 24.49 & \textbf{18.44} & 21.02 & 23.05 & 16.36 & 23.78 & 20.73 & 3.24 \\
propagate\_XGBoost & \textbf{15.68} & \textbf{27.69} & 32.04 & 18.24 & \textbf{22.85} & \textbf{23.80} & \textbf{18.05} & \textbf{28.92} & \textbf{23.41} & 5.46 \\
\hline
HT    &  4.67 & 10.03 & 15.27 &  7.20 &  6.43 &  4.10 &  6.11 & 16.63 &  8.80 & 4.46 \\
HAT               &  1.49 &  0.55 &  1.77 &  1.36 &  0.87 &  1.10 &  2.48 &  5.99 &  1.95 & 1.62 \\
LR                &  6.91 &  9.11 & 12.25 &  6.74 &  7.32 &  5.77 &  7.52 & 15.79 &  8.93 & 3.19 \\
static\_DT         &  7.39 & 10.88 & 12.01 &  4.98 &  4.71 &  4.31 &  4.13 &  6.23 &  6.83 & 2.86 \\
r\_DT              &  7.39 & 11.31 &  8.07 &  7.05 &  7.52 &  8.13 &  7.62 &  6.59 &  7.96 & \textbf{1.35} \\
propagate\_DT      &  7.39 & 13.82 & 14.68 & 10.72 & 10.74 &  9.15 &  8.63 & 19.83 & 11.87 & 3.80 \\
static\_TabSRA     & 12.60 & 17.91 & 23.69 & 11.06 & 13.62 &  8.91 &  9.81 & 20.94 & 14.82 & 5.08 \\
r\_TabSRA          & 12.60 & 14.17 & 22.68 & 15.03 & 15.21 & 13.95 & 11.47 & 20.19 & 15.66 & 3.58 \\
u\_TabSRA          & 12.60 & 19.10 & 26.38 & 16.65 & 18.73 & 19.01 & 13.99 & 19.56 & 18.25 & 3.91 \\
propagate\_TabSRA  & 12.60 & 19.24 & \textit{26.74} & 15.52 & \textit{18.88} & 20.56 & 14.41 & 22.86 & 18.85 & 4.35 \\
static\_EBM        & \textit{13.36} & 20.80 & 25.94 & 12.26 & 14.85 & 16.73 &  9.27 & 14.24 & 15.93 & 4.91 \\
r\_EBM             &  \textit{13.36} & 22.03 & 22.92 & 15.33 & 16.85 & 18.06 & 12.72 & 17.49 & 17.35 & 3.45 \\
B\_EBM             &  \textit{13.36} & 20.80 & 25.94 & \textit{17.21} & 17.55 & 21.55 & 15.06 & 22.06 & 19.04 & 3.89 \\
propagate\_EBM     &  \textit{13.36} & \textit{22.88} & 24.92 & 16.72 & 17.44 & \textit{21.59} & \textit{15.14} & \textit{26.20} & \textit{19.78} & 4.45 \\
\bottomrule
\end{tabular}
\end{center}
\vspace{-5mm}
\end{table*}
}

As shown in Table \ref{tab:fraudataset}, the best batch solution, XGBoost, consistently outperforms (over the 8 months) the best instance incremental solution, namely Adaptive Random Forest (ARF). Similarly, the Explainable Boosting Machine (EBM) provides the best average AUCPR among the inherently interpretable solutions, surpassing all the instance incremental solutions. Besides the overall superior predictive performance of batch incremental solutions in this problem, the \textbf{propagate} strategy, which consists of using the tree last months (chunks) of labeled data, tends to provide the best results. For XGBoost, the average AUCPR gain is up to 22\% compared to the retrain strategy (r\_XGBoost) and 12\% compared to the stacking strategy. 
We argue that this because the target event (fraudulent examples) is rare. Consequently, using only one chunk (one month) of observation should not be enough to retrain a consistent learning model, and finding the best weighting combination for the stacking approach is not an easy task. This finding is following \cite{wang2003mining} where the author demonstrates that varying the number of observations in the chunks (from 3,000 to 12,000 observations) increases linearly the benefit (the number of frauds detected) by the learning model. Particularly for the fraud dataset investigated in our study, the number of fraudulent examples seen during the training or the updating of the model is very crucial for the predictive performance, highlighted by the fact that a static approach with 3 months of training data (static\_XGBoost) outperforms in average the periodical retraining with 1 month (chunk) of observations (r\_XGBoost). 
It also indicates that some concepts may reappear over time (e.g., static\_XGBoost outperforms propagate\_XGBoost for March). Overall, the results on this dataset suggest that storing some past observations may be advantageous for the streaming learning model.\\
\indent We also note an intriguing result (Table \ref{tab:fraudataset})  for inherently interpretable instance incremental models: Hoeffding Tree (HT) outperforms the HT equipped with the ADWIN drift detector (HAT) in contrast to the results on the \textit{Generated benchmark} (Table \ref{tab:delay0}). We argue that for such a dataset where the target event of interest is very rare (highly imbalanced dataset), explicit drift handling may be biased toward negative examples. This because many statistical drift detectors rely on accuracy metrics or error rates.  
Consequently, the HAT model may discard some parts (or branches) of the tree that are specialized in identifying concepts (fraudulent schemes) that may reoccur over time.
\section{Conclusion and Discussions}
We proposed a thorough comparison between batch and instance incremental models by examining the usefulness of the latter in situations where labels become available 
with varying delays, and interpretability is crucial. 
Our findings on commonly used synthetic data streams demonstrate that the best batch incremental approach (fully complex or inherently interpretable), combined with an effective updating strategy, provides similar or better performance than instance incremental counterparts. Additionally, the superior performance of batch learners is illustrated using a real-world fraud detection problem where the target class is scarce. We explain why, for such a dataset, it is necessary to store some observations while waiting for the labels due to the delay. Furthermore, we demonstrate how storing some old but limited chunks of observations can benefit batch incremental models, leading to up to a 22\% relative improvement in AUCPR compared to using only one chunk.
Although we illustrate the usefulness of batch incremental solutions, we do not cover the choice of the optimal batch (chunk) size, as it may depend on several factors (e.g., the frequency of the target event, bias handling for underrepresented groups) that we leave for future study. Nevertheless, we believe that choosing the learning batch based on evaluation date (daily, monthly) can be a good starting point, as illustrated with the fraud dataset.\\
\indent We hope our findings will encourage and guide researchers in developing new learning strategies for evolving data streams, bridging the gap between academic research and real-world business use cases.
\bibliographystyle{splncs04}
\bibliography{mybibliography}
\end{document}